\documentclass{article}

\PassOptionsToPackage{numbers, compress}{natbib}

\usepackage[main, final]{neurips_2026}

\usepackage[table,x11names]{xcolor}
\usepackage{graphicx}       
\usepackage{multirow}       
\usepackage{makecell}       
\usepackage{amsmath}        
\usepackage{bbm}            
\usepackage{siunitx}        
\usepackage{subcaption}     
\usepackage{kotex}          

\usepackage[utf8]{inputenc} 
\usepackage[T1]{fontenc}    
\usepackage{hyperref}       
\usepackage{url}            
\usepackage{booktabs}       
\usepackage{amsfonts}       
\usepackage{nicefrac}       
\usepackage{microtype}      
\usepackage{xcolor}         

\newcommand{\ie}{\textit{i}.\textit{e}.}

\title{Disambiguating 2D-3D Correspondences in Gaussian Splatting-based Feature Fields for Visual Localization}

\author{
  Miso Lee\;\;\;\; Sangeek Hyun\;\;\;\; Yerim Jeon\;\;\;\; Jae-Pil Heo\thanks{Corresponding author} \\[2pt]
  Sungkyunkwan University
}

\begin{document}
\maketitle

\begin{abstract}
While Gaussian Splatting-based Feature Fields (GSFFs) have shown promise for visual localization, this paper highlights that photometrically optimized GSFFs are inherently not suitable for 2D-3D matching.
The volumetric extent of each Gaussian induces many-to-one pixel-to-point mappings that destabilize PnP-based pose estimation, while photometric optimization gives rise to superfluous Gaussians devoid of multi-view consistency.
To address these issues, we propose \textbf{SplitGS-Loc}, a GSFFs construction framework for localization, which disambiguates 2D-3D correspondences by exploiting Gaussian attributes.
Our key design, Mixture-of-Gaussians-based splitting, decomposes each Gaussian into smaller Gaussians, replacing ambiguous many-to-one with precise one-to-one correspondences.
In parallel, we exploit composition weights from GS rasterization to select Gaussians that significantly and consistently contribute across multiple views and aggregate discriminative features through strong pixel-Gaussian associations.
The resulting compact yet discriminative feature fields enable stable PnP convergence, achieving competitive performance on localization benchmarks. Extensive experiments validate that SplitGS-Loc extends the utility of photometric GSFFs to accurate and efficient localization by exploiting Gaussian attributes, without per-scene training for feature fields or iterative pose refinement.
The code is available at: \url{https://github.com/misoleekr/SplitGS-Loc}.
\end{abstract}    
\section{Introduction}
\label{sec:intro}

Visual localization is a fundamental computer vision task that estimates the camera pose from an image, providing the foundation for autonomous driving, robotics, and AR/VR systems.
When a predefined 3D map is given, the camera pose can be obtained by matching 2D pixels with 3D points and using Perspective-n-Points (PnP)~\cite{gao2003pnp} with RANSAC~\cite{fischler1981ransac}.
Nowadays, as more expressive 3D representations like Gaussian Splatting (GS)~\cite{kerbl20233dgs} and 3D feature fields~\cite{kerr2023lerf, zhou2024feature3dgs, qin2024langsplat} have been actively explored, localization performance has been further advanced with GS-based Feature Fields (GSFFs)~\cite{sidorov2025gsplatloc, pietrantoni2025gsffs, huang2025stdloc}.
GSFFs are usually constructed by adding features to Gaussians with feature rendering loss~\cite{qin2024langsplat, sidorov2025gsplatloc, pietrantoni2025gsffs},
or jointly optimizing colors and features~\cite{zhou2024feature3dgs, huang2025stdloc}.
This allows camera pose refinement through 2D-2D matching between query image features and rendered image features.

Recently, STDLoc~\cite{huang2025stdloc} enables direct 2D-3D matching by sampling both query image features and GSFFs, reducing the computational overhead during inference.
However, it demands hours of multi-stage optimization per scene;
even after optimizing GSFFs, a scene-specific detector is required to align 2D keypoints with sampled Gaussians as GSFFs alone are not suitable for 2D-3D matching.
Moreover, to achieve high localization accuracy, it further stores the full GSFFs and performs iterative dense feature-based pose refinement.
Thus, although STDLoc establishes an effective sparse-to-dense localization pipeline, it still relies on additional query-side detection and dense 2D-2D refinement.

\begin{figure}[t]
\centering
\includegraphics[width=1.0\linewidth]{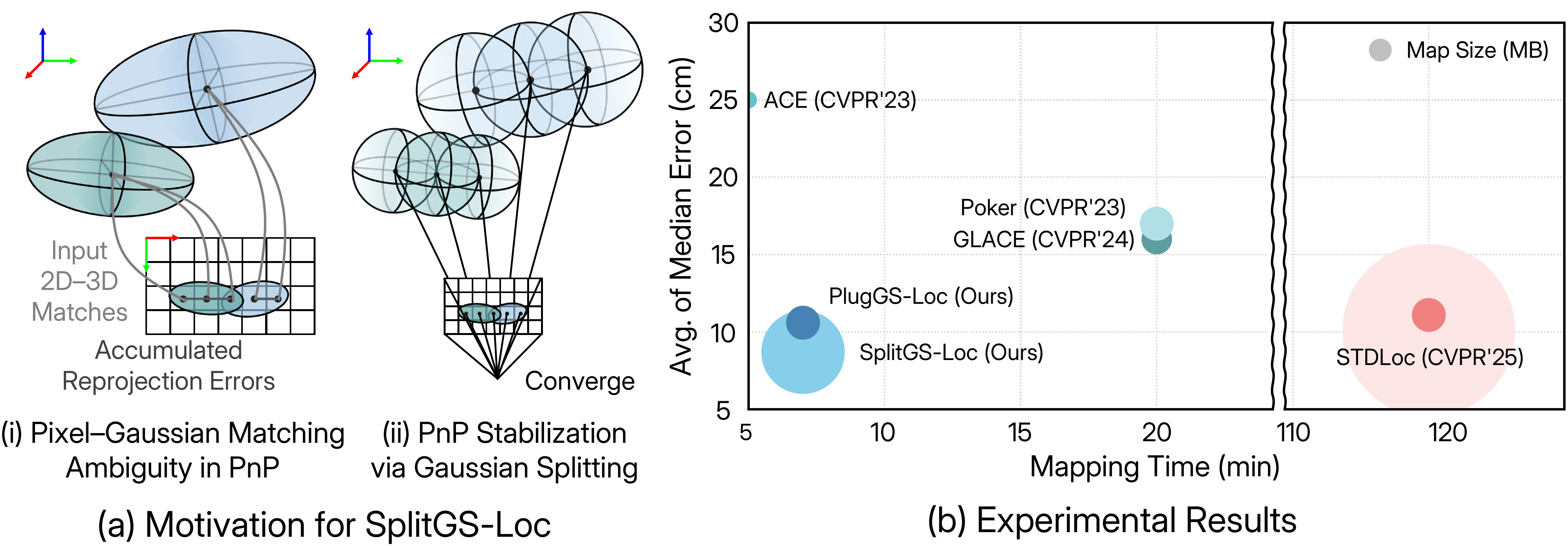}
\caption{
\textbf{Motivation and experimental results.}
(a) Each 3D Gaussian covers a volumetric region and can thus correspond to multiple image pixels.
However, PnP takes 2D-3D point correspondences as input; when each Gaussian is approximated by a single 3D point (e.g., its center), this leads to many-to-one correspondences.
We resolve this by decomposing Gaussians into smaller ones, increasing valid one-to-one correspondences and stabilizing PnP-based localization.
(b) The figure illustrates the trade-off between mapping efficiency and localization accuracy on Cambridge Landmarks~\cite{kendall2015cambridge}. 
While recent GSFFs-based methods achieve superior accuracy but sacrifice mapping efficiency, whereas our proposed frameworks attain a favorable balance by disambiguating pixel-Gaussian correspondences.
}
\label{fig:motivation}
\end{figure} 
We discover that this inefficiency can be resolved by directly addressing the photometrically optimized Gaussians for 2D-3D matching.
First, we reveal that the volumetric extent of Gaussians entails a structural mismatch with PnP, which requires strict point-to-point correspondences as input.
As illustrated in Fig.~\ref{fig:motivation} (a), when multiple 2D pixels correspond to the same Gaussian and are all mapped to a single 3D point, these many-to-one correspondences lead to geometric inconsistency.
Consequently, PnP with RANSAC becomes unstable and ultimately outputs a suboptimal pose.
In addition, numerous redundant Gaussians are spawned to capture view-dependent photometric effects~\cite{kerbl20233dgs, yang2024specgaussian, hanson2025pup}.
Hence, they serve no meaningful role in 2D-3D matching under novel viewpoints due to the lack of multi-view consistency.

To this end, we propose a localization-specialized GSFFs construction framework, \textbf{SplitGS-Loc}, which disambiguates pixel-Gaussian correspondences by utilizing Gaussian attributes.
Central to our approach is Mixture-of-Gaussians-based splitting, which decomposes each Gaussian along its major axis.
This fast, deterministic, and feature-independent decomposition yields fine-grained Gaussians that serve as the foundation for stable 2D-3D matching.
Yet, for those Gaussians to produce valid one-to-one matches at inference, each must be paired with its own distinctive feature.

This motivates us to devise a direct feature lifting pipeline, PlugGS-Loc, which exploits composition weights derived from the rasterization process.
Specifically, given training views and photometric GS, it proceeds to identify salient Gaussians that significantly and consistently contribute to pixel rasterization across multiple views, 
and then assign to each selected Gaussian a pixel-level feature aggregated through strong pixel-Gaussian associations.  
This prevents feature blurring due to rendering-based optimization~\cite{wu2024opengaussian, jun2025drsplat} and promotes multi-view consistency.

Altogether, SplitGS-Loc disambiguates 2D-3D correspondences in GSFFs by alleviating many-to-one matching ambiguity through Gaussian decomposition and assigning a distinctive feature to each split Gaussian.
As a result, we achieve competitive performance across various indoor and outdoor scenes~\cite{kendall2015cambridge, shotton20137scenes}.
Extensive experiments demonstrate that reducing many-to-one pixel-Gaussian matches leads to stable PnP convergence, yielding faster runtime and better accuracy.
From a broader perspective, this work reframes GSFFs from exhaustive 3D maps to deployable ones, restoring the balance between performance and practicality as illustrated in Fig.~\ref{fig:motivation} (b).
\section{Related Work}
\label{sec:related_work}

\noindent{\textbf{Classical methods}} follow a multi-stage pipeline~\cite{li2010location, sattler2016AS, sattler2017large3d, sarlin2019hloc, humenberger2020r2d2, sarlin2021pixloc, giang2024DeViLoc}.
Given a query image, (i) similar reference views are first retrieved from an image database~\cite{torii2015densevlad, arandjelovic2016netvlad}, 
(ii) local feature extraction and matching~\cite{lowe2004SIFT, sarlin2020superglue, detone2018superpoint} are performed between the query and retrieved images, 
(iii) the resulting 2D-2D correspondences are lifted to 2D-3D correspondences, as each reference image is registered to a 3D map via Structure-from-Motion (SfM), 
and (iv) the camera pose is estimated using the Perspective-n-Point (PnP)~\cite{gao2003pnp} algorithm within a RANSAC loop~\cite{fischler1981ransac}.
Although this pipeline provides strong performance, it relies on a large-scale image database and 3D map, together with multiple stages.

\noindent{\textbf{Deep learning-based methods}} have been developed in response to the need for efficiency in practical applications, aiming toward more streamlined pipelines.
Scene Coordinate Regression (SCR) methods~\cite{brachmann2018dsac++, brachmann2021dsacstar, brachmann2023ace, wang2024glace} train a network to predict a 3D scene coordinate for each pixel, thereby enabling PnP-based pose estimation without an image database or SfM map.
Several approaches~\cite{kendall2015cambridge, kendall2017posenet17, brahmbhatt2018mapnet} further accelerate the inference stage by directly regressing the camera pose from a 2D image, making them independent of the PnP+RANSAC stage.
While these Absolute Pose Regression (APR) methods achieve fast inference, they trade localization accuracy for inference speed and require scene-specific training, thus limiting their scalability to real-world multi-scene applications~\cite{sattler2019understanding, brachmann2023ace, dong2025reloc3r}.
Recent works have attempted to overcome these limitations by accelerating per-scene training to a matter of minutes~\cite{brachmann2023ace, wang2024glace, chen2024marepo} or by optimizing a single model that can handle multiple scenes~\cite{shavit2023c2f-mst, xie2023ufvl, lee2024actmst}.

\noindent{\textbf{NeRF-based methods}} were initially introduced to bridge the gap between efficiency and accuracy in APR,
either by augmenting training data via novel view synthesis~\cite{chen2021directpn, moreau2022lens} or by aligning query features with implicit features~\cite{chen2022dfnet, moreau2023crossfire} from NeRF~\cite{mildenhall2020nerf, martin2021nerf-w, muller2022instantngp}.
Beyond utilizing it for APR, several works leverage rendered depth and features from NeRF to reintroduce PnP-based pose estimation, achieving strong accuracy with compact maps~\cite{moreau2023crossfire, zhao2024pnerfloc, zhou2024nerfmatch}.

\noindent{\textbf{GS-based methods}} have further advanced localization accuracy upon explicit 3D Gaussian representations.
Leveraging the real-time, high-fidelity photometric rendering of GS, fast and robust pose refinement has become readily available~\cite{liu2025gscpr, khatib2025gsvisloc, Li2025rap, wang2025GSRelocNet}.
In parallel, as 3D feature fields~\cite{kerr2023lerf, qin2024langsplat, zhou2024feature3dgs} embed preprocessed 3D features aligned with 2D ones, GS-based feature fields (GSFFs) naturally support feature rendering-based pose refinement.
Notably, STDLoc~\cite{huang2025stdloc} treats GSFFs as the predefined 3D map itself, employing Gaussians not only for feature rendering but directly for 2D-3D feature matching, thereby establishing an end-to-end localization pipeline upon GSFFs.

However, this entails hours of per-scene training for feature fields, and the resulting feature fields are not inherently suited for 2D-3D matching.
We attribute this to redundant Gaussians and ambiguous pixel-Gaussian correspondences induced by photometric optimization, both undermining direct 2D-3D matching.
In contrast, we construct localization-specialized GSFFs through composition weight-based feature lifting and Mixture-of-Gaussians-based splitting, explicitly decomposing Gaussian primitives to alleviate many-to-one matching ambiguity without per-scene training for feature fields.
\section{Method}
\label{sec:method}

\subsection{Preliminary}
\label{sec:preliminary_gsff}
3D Gaussian Splatting (3DGS)~\cite{kerbl20233dgs} extends 3D point clouds into an explicit set of 3D Gaussian primitives $\mathcal{G}=\{g_n\}_{n=1}^{N_g}$, where $N_g$ denotes the number of Gaussians, supporting efficient differentiable volumetric rendering.
Each 3D Gaussian is parameterized by a mean $\mu$, an anisotropic covariance $\Sigma = R S S^\top R^{\top}$ (with $R$ and $S$ representing rotation and scale, respectively), an opacity $\alpha$, and spherical harmonic (SH) color coefficients $\mathbf{c}$.
For the $j$-th pixel in $i$-th image $I_{ij}$, the rendered color is computed as $\hat{I}_{ij} = \sum_n w(g_n,I_{ij}) \, \mathbf{c}_n$, 
where $w(g_n,I_{ij})$ denotes the weight assigned to each Gaussian during the rasterization process, reflecting both the accumulated transmittance along the ray and its opacity.
We hereafter refer to this term as the composition weight.

Building upon the photometric 3DGS framework, GS-based Feature Fields (GSFFs)~\cite{zhou2024feature3dgs, qin2024langsplat} augment each Gaussian $g_n$ with a feature vector $\mathbf{z}_n$.
For localization, GSFFs serve as a 3D map in which each Gaussian acts as a 3D point with $\mathbf{z}_n$ as its descriptor.
The camera pose of a query image is then estimated by PnP+RANSAC~\cite{gao2003pnp, fischler1981ransac} on 2D-3D matches obtained via feature similarity.
To construct such GSFFs, a common formulation is to optimize $\mathbf{z}_n$ by supervising the rendered feature map $\hat{\mathbf{f}}_{ij} = \sum_n w(g_n,I_{ij})\,\mathbf{z}_n$ against the corresponding pixel-level features $\mathbf{f}_{ij}$ extracted by a 2D image encoder, so that the resulting feature field is aligned with the encoder's feature space.
However, we argue that these approaches are less effective for direct matching, 
as rendering-based optimization causes feature blurring by distributing gradients over hundreds of Gaussians per ray~\cite{wu2024opengaussian, jun2025drsplat}, thereby weakening 2D-3D correspondences for PnP.
Accordingly, we adopt a direct feature lifting pipeline inspired by recent open-vocabulary semantic segmentation works~\cite{jun2025drsplat, marrie2025ludvig}, 
but tailor it for localization.

\subsection{PlugGS-Loc}
\label{sec:method_pluggs-loc}
\begin{figure*}[t]
\centering
\includegraphics[width=1.0\linewidth]{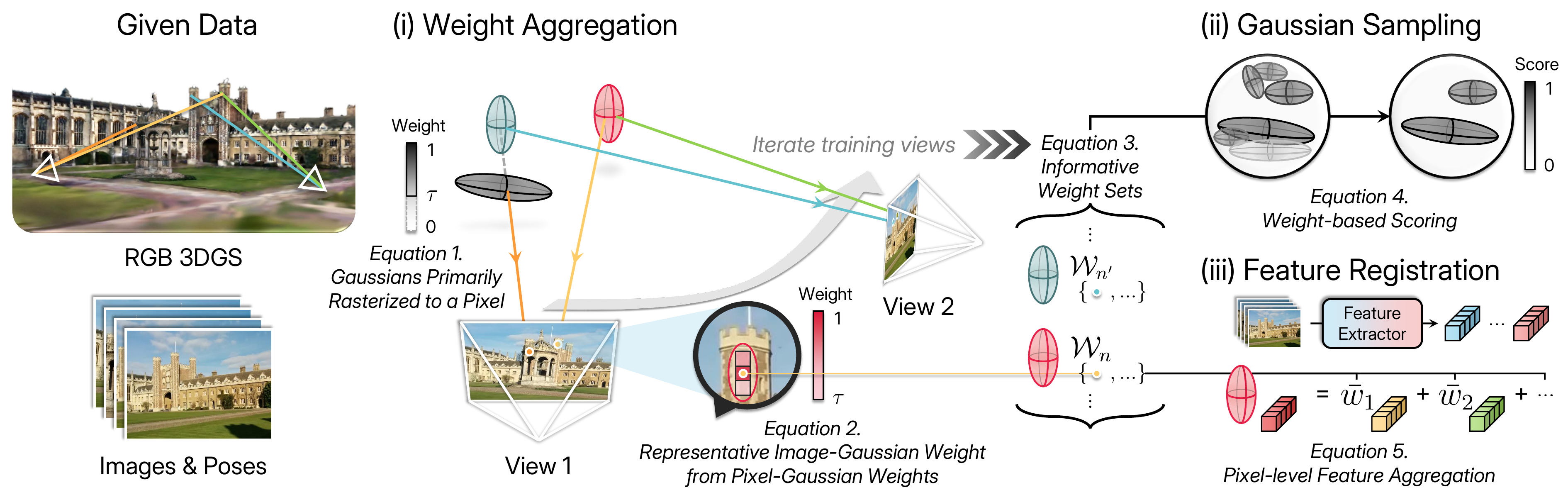}
\caption{
\textbf{PlugGS-Loc pipeline.} 
The figure illustrates how to construct GSFFs for localization when pretrained RGB 3DGS and training views are given.
(i) For each view $I_i$, we render and identify Gaussians primarily rasterized above the weight threshold $\tau$ in Eq.~\ref{eq:primary_gaussians} for each pixel, and select a representative image-Gaussian weight among them for memory efficiency as in Eq.~\ref{eq:image_gaussian_weight}.
After iterating over all training views, the resulting informative weight sets defined in Eq.~\ref{eq:informative_weight_set} serve as the core of both (ii) Gaussian Sampling and (iii) Feature Registration.
The former selects essential Gaussians, while the latter assigns multi-view aggregated features to those Gaussians.
Note that the entire process is streamlined and thus can be completed within a minute per scene.
}
\label{fig:pluggs-loc}
\end{figure*}

\paragraph{Weight aggregation.}
To identify Gaussians that are suitable for 2D-3D matching, we first gather the set of Gaussians primarily rasterized to the corresponding pixel $I_{ij}$ by applying a threshold $\tau$ on their composition weights:
\begin{equation}
    \hat{\mathcal{G}}_{ij} = \{ g_n \in \mathcal{G} \mid w(g_n, I_{ij}) \ge \tau \}.
    \label{eq:primary_gaussians}
\end{equation}
Intuitively, this preserves Gaussians that contribute more than $\tau$ (e.g., $\tau=0.1$ $\rightarrow$ 10\%) of the total composition weight for the pixel $I_{ij}$.
However, storing all of them requires a huge amount of memory, which is infeasible in practice.
Instead, we exploit only the maximum composition weight over all pixels in the view, forming one representative weight for each image-Gaussian pair.
For each Gaussian $g_n$ and training image $I_i$, it is defined as:
\begin{equation}
    \tilde{w}(g_n,I_i) :=
    \begin{cases}
    \max\limits_{j} \big( w(g_n,I_{ij})\big), & \text{if } \cup \hat{\mathcal{G}}_{ij} \neq \emptyset,\\
    \, \, 0, & \text{otherwise.}
    \end{cases}
    \label{eq:image_gaussian_weight}
\end{equation}
In essence, $\tilde{w}(g_n,I_i)$ captures the strongest pixel-Gaussian contribution in a view $I_i$ as illustrated in Fig.~\ref{fig:pluggs-loc}.
We then iterate all training images to collect multi-view contributions,
finally obtaining a compact and informative weight set for each Gaussian as:
\begin{equation}
    \mathcal{W}_n = \{ \tilde{w}(g_n,I_i) \mid \tilde{w}(g_n,I_i) > 0 \}.
    \label{eq:informative_weight_set}
\end{equation}
$\{ \mathcal{W}_n \}_{n=1}^{N_g}$ lay the foundation for both Gaussian sampling and feature registration, supporting the streamlined GSFFs construction pipeline.
Note that only non-zero weights are considered, so Gaussians with $\mathcal{W}_n=\emptyset$ are excluded in the next step.

\paragraph{Gaussian sampling.}
As exhaustive 2D-3D matching becomes computationally expensive with numerous Gaussians (and as many of them are redundant due to photometric optimization), it is essential to prune superfluous ones.
Following STDLoc~\cite{huang2025stdloc}, we define each local 3D region by randomly sampling anchor Gaussians to cover the scene and grouping their k-nearest neighbors in terms of the L2 distance between Gaussian centers.
Then, we score each Gaussian's importance and retain only the one with the highest score within each local region.
Unlike STDLoc~\cite{huang2025stdloc}, we utilize the informative weight set $\mathcal{W}_n$ to measure the importance of each Gaussian as
\begin{equation}
    \psi (g_n) = \frac{1}{|\mathcal{W}_n|}\sum_{\tilde{w} \in \mathcal{W}_n} \tilde{w},
    \label{eq:gussian_score}
\end{equation}
where $|\mathcal{W}_n|$ denotes the cardinality of $\mathcal{W}_n$.
This weight-based criterion favors Gaussians with high average contributions over views with non-zero weights.
Since it measures the average contribution conditioned on visibility, a Gaussian strongly observed in only a few views could dominate one moderately observed in many views.
Nevertheless, under dense training views, this scoring and sampling within local regions empirically yield multi-view supported Gaussians distributed across the scene, supporting robust matching in novel views while maintaining a compact map.

\paragraph{Feature registration.}
Instead of rendering-based optimized $\mathbf{z}_n$, we adopt $\bar{\mathbf{z}}_n$, which is computed as a weighted sum of the corresponding pixel features as follows:
\begin{equation}
    \bar{\mathbf{z}}_n = \sum_{(i,j^*) \in \mathcal{P}_n} \bar{w}_{ij^*} \bar{\mathbf{f}}_{ij^*},
\label{eq:feature_aggregation}
\end{equation}
where $\mathcal{P}_n = \{(i,j^*) \mid w(g_n,I_{ij^*}) = \tilde{w}(g_n,I_i)\}$ is a set of pixel indices that correspond to $\mathcal{W}_n$,
$\bar{w}$ denotes the softmax-normalized weight within $\mathcal{W}_n$, 
and $\bar{\mathbf{f}}$ denotes pixel-level image feature $\mathbf{f}$ normalized along the channel dimension.
It enables us to obtain multi-view aggregated representations, 
finally constructing GSFFs within a minute. 
In summary, PlugGS-Loc efficiently identifies the most informative Gaussians and associates them with features exhibiting strong pixel-Gaussian correspondence, resulting in robust 2D-3D matching for localization.

\subsection{SplitGS-Loc}
\label{sec:method_splitgs-loc}
SplitGS-Loc is built on the key insight that photometrically optimized Gaussians are ill-suited for 2D-3D matching since their volumetric extent induces a structural mismatch with PnP.
A Gaussian can be associated with several pixels, but since PnP requires point-to-point correspondences as input, many-to-one matches inevitably accumulate reprojection errors and thus produce suboptimal results.
To address the issue, we introduce a Mixture-of-Gaussians-based splitting strategy that decomposes each Gaussian along its major axis, the direction most prone to ambiguity.

\paragraph{Mixture of Gaussians and 3DGS.}
A Gaussian distribution, omitting the index $n$ for notational simplicity, can be formulated as $p(x) \sim \mathcal{N}(\mu,\Sigma)$,
which extends to a mixture as
\begin{equation}
    p(x) = \sum_{k=1}^{K}\lambda_k\mathcal{N}(x|\mu_k,\Sigma_k),
\end{equation}
where the components form a linear combination with coefficients $\lambda_k$ satisfying $\sum_k \lambda_k = 1$.
In 3DGS, this corresponds to distributing a Gaussian’s opacity $\alpha$ across its components such that the $k$-th child carries opacity $\lambda_k \alpha$.
The covariance, as introduced in Sec.~\ref{sec:preliminary_gsff}, is given by $\Sigma = R S S^{\top} R^{\top}$,
with the scale matrix $S=\mathrm{diag}(s_x,s_y,s_z)$ describing the radii of the Gaussian ellipsoid along the $x$-, $y$-, and $z$-axes, respectively.
Since a multivariate Gaussian factorizes into independent components along its principal axes, we hereafter consider a mixture of 1D Gaussians.

\paragraph{Mixture-of-Gaussians-based splitting.}
In order to reduce the pixel-Gaussian matching ambiguity, particularly along the major direction with the highest ambiguity tendency, we decompose each 3D Gaussian by symmetrically adding two side components along its major axis.
Formally, in the canonical space and considering only the major axis $x$ such that $\tilde{p}(x)\sim\mathcal{N}(0,s^2)$,
$\tilde{p}(x)$ can be represented as a mixture of Gaussians:
\begin{equation}
    \tilde{p}(x) \approx \tilde{q}(x) = \lambda q_{-}(x) + \lambda_\text{o} q_\text{o}(x) + \lambda q_{+}(x),
    \label{eq:split_definition}
\end{equation}
where $s$ denotes the standard deviation (\ie, scale) along the selected major axis, $2\lambda+\lambda_\text{o}=1$, and each component is defined as:
\begin{equation}
    \begin{aligned}
    q_{-}(x) &\sim \mathcal{N}(-\beta s,\ \sigma^2),\\
    q_\text{o}(x) &\sim \mathcal{N}(0,\ \sigma_\text{o}^2),\\
    q_{+}(x) &\sim \mathcal{N}(+\beta s,\ \sigma^2),
\end{aligned}
\label{eq:child_definition}
\end{equation}
where $\beta>0$ controls the side offsets, and $\sigma^2,\sigma_\text{o}^2$ are the component variances. 

Based on this formulation, the parameters are derived by matching the first four moments of $\tilde{p}(x)$ and $\tilde{q}(x)$.
Moment matching determines $\lambda_\text{o}=\frac{2}{3}$, $\lambda=\frac{1}{6}$, and $\sigma^2=\sigma_\text{o}^2=(1-\frac{\beta^2}{3})s^2$ for any $\beta \in (0, \sqrt{3})$.
Under this configuration, the side components are positioned along the ellipsoid's major axis at $\pm \beta s$, spanning a wider extent of the original distribution and thus reducing the pixel-Gaussian matching ambiguity.
For the remaining axes, the child Gaussians follow the original distribution, and are finally transformed back to the original coordinate frame.
Note that this process runs with one million Gaussians in under one second.
A detailed derivation is provided in the appendix.

\paragraph{Integration of splitting strategy and PlugGS-Loc.}
Once splitting strategy is applied, the resulting child Gaussians are seamlessly integrated into the PlugGS-Loc to transform dense GS into localization-specialized GSFFs.
Regarding Gaussian sampling, however, directly applying Eq.~\ref{eq:gussian_score} to child Gaussians would undermine the one-to-one correspondences established by splitting.
This is because multiple child Gaussians within the same local region would collapse into a single representative.
To prevent this, we temporarily aggregate the child Gaussians only at the sampling stage, applying the new score aggregated into the original Gaussian, defined as:
\begin{equation}
    \bar{\psi}(g_n) = \frac{1}{3}\bigl(\psi(g_{n_-}) + \psi(g_{n_\text{o}}) + \psi(g_{n_+})\bigr),
\label{eq:gaussian_score_split}
\end{equation}
where $g_{n_-}$, $g_{n_\text{o}}$, and $g_{n_+}$ are child Gaussians split from the original Gaussian $g_{n}$.
Then, sampling and scoring are performed based on the original Gaussians’ centers $\{\mu_{n}\}_{n=1}^{N_g}$ and aggregated scores $\{\bar{\psi}(g_n)\}_{n=1}^{N_g}$, and corresponding child Gaussians of the selected parents are finally preserved.
Subsequently, they are independently paired with different pixel features during feature registration; therefore, the salient region once represented by a single descriptor is described by multiple fine-grained ones.
As a result, SplitGS-Loc yields compact yet more discriminative GSFFs, mitigating the pixel-Gaussian matching ambiguity, thus enabling accurate matching and stable PnP convergence.
\section{Experiment}
\label{sec:experiment}
\begin{table}[t]
\caption{\textbf{Localization results on Cambridge Landmarks~\cite{kendall2015cambridge}.}
We report median translation and rotation errors.
For all GS-based methods, mapping time includes RGB 3DGS~\cite{kerbl20233dgs} training time (about 6min),
and map size includes per-scene optimized networks and feature fields on top of 3DGS ($\sim$80MB).
$^+$ denotes the version with iterative camera pose refinement using GS-CPR~\cite{liu2025gscpr}, and $^\dagger$ denotes methods that need additional training to adapt the representation for localization per scene.}
\centering
\resizebox{\textwidth}{!}{
\begin{tabular}{clcccccccccccc}
    \toprule
    \multicolumn{1}{l}{} 
    & 
    & \makecell{Mapping\\Time}
    & \makecell{Pose\\Refine}
    & \makecell{Size\\(MB)}
    & \makecell{Great\\Court}
    & \makecell{King's\\College}
    & \makecell{Old\\Hospital}
    & \makecell{Shop\\Facade}
    & \makecell{St. M.\\Church}
    & \makecell{Average\\(cm/$^\circ$)}
    \\ 
    \midrule
    \multirow{3}{*}{\rotatebox{90}{FM}}
    & AS (SIFT)~\cite{sattler2016AS}
    & 0 & & $\sim$200
    & 24/0.13 & 13/0.22 & 20/0.36 & 4/0.21 & 8/0.25 & 13.8/0.23 \\
    & HLoc (SP+SG)~\cite{sarlin2019hloc, sarlin2020superglue}
    & 0 & & $\sim$800
    & 17.7/0.11 & 11.0/0.20 & 15.1/0.31 & 4.2/0.20 & 7.0/0.22 & 11.0/0.21 \\
    & PixLoc~\cite{sarlin2021pixloc}
    & 0 & \checkmark & $\sim$600
    & 30/0.14 & 14/0.24 & 16/0.32 & 5/0.23 & 10/0.34 & 15.0/0.25 \\
    \midrule
    \multirow{3}{*}{\rotatebox{90}{SCR}}
    & DSAC*~\cite{brachmann2021dsacstar}
    & 15h & & 28
    & 34/0.2 & 18/0.3 & 21/0.4 & 5/0.3 & 15/0.6 & 19/0.4 \\
    & ACE~\cite{brachmann2023ace}
    & 5min & & 4
    & 43/0.2 & 28/0.4 & 31/0.6 & 5/0.3 & 18/0.6 & 25/0.4 \\
    & GLACE~\cite{wang2024glace}
    & 20min & & 13
    & 19/0.1 & 19/0.3 & 17/0.4 & 4/0.2 & 9/0.3 & 14/0.3 \\
    \midrule
    \midrule
    \multicolumn{11}{l}{Neural Scene Representation$^\dagger$} \\
    \midrule
    \multirow{3}{*}{\rotatebox{90}{NeRF}}
    & CROSSFIRE~\cite{moreau2023crossfire}
    & 15h & \checkmark & 50
    & N/A & 47/0.7 & 43/0.7 & 20/1.2 & 39/1.4 & N/A \\
    & PNeRFLoc~\cite{zhao2024pnerfloc}
    & hrs & \checkmark & $\sim$50
    & 81/0.25 & 24/0.29 & 28/0.37 & 6/0.27 & 40/0.55 & 35.8/0.35 \\
    & NeRFMatch~\cite{zhou2024nerfmatch}
    & hrs & \checkmark & $\sim$55
    & 19.6/0.09 & 12.5/0.23 & 20.9/0.38 & 8.4/0.40 & 10.9/0.35 & 14.5/0.29 \\
    \midrule
    \multirow{8}{*}{\rotatebox{90}{GS}}
    & GSplatLoc~\cite{sidorov2025gsplatloc}
    & hrs & \checkmark & $>$100
    & N/A & 31/0.49 & 16/0.68 & 4/0.34 & 14/0.42 & N/A \\
    & GSFFs-PR~\cite{pietrantoni2025gsffs}
    & hrs & \checkmark & $>$100
    & N/A & 17/0.26 & 18/0.36 & 4/0.25 & 8/0.26 & N/A \\
    & STDLoc~\cite{huang2025stdloc}
    & 2h & & 16
    & \textbf{13.1}/0.08 & 15.2/0.17 & 18.6/0.36 & 3.5/0.17 & 5.3/0.19 & 11.1/0.19 \\
    & STDLoc (PR)~\cite{huang2025stdloc}
    & 2h & \checkmark & 415
    & 15.7/\textbf{0.06} & 15.0/0.17 & 11.9/0.21 & \textbf{3.0/0.13} & \textbf{4.7}/0.14 & \textbf{10.1/0.14} \\
    & GSVisLoc$^+$~\cite{liu2025gscpr, khatib2025gsvisloc}
    & hrs & \checkmark & $\sim$80
    & N/A & 23/0.3 & 22/0.42 & 8/0.29 & 14/0.45 & N/A \\
    & RAP$^+$~\cite{liu2025gscpr, Li2025rap}
    & hrs & \checkmark & $\sim$80
    & 22/0.15 & 15/0.23 & 18/0.38 & 5/0.23 & 9/0.23 & 13.8/0.24 \\
    & GS-RelocNet~\cite{wang2025GSRelocNet}
    & hrs & & $>$100
    & N/A & 11/0.19 & 13/0.26 & 4/0.18 & 7/0.15 & N/A \\
    & GS-RelocNet (PR)~\cite{wang2025GSRelocNet}
    & hrs & \checkmark & $>$100
    & N/A & \textbf{9/0.15} & \textbf{10/0.19} & \textbf{3}/0.15 & 5/\textbf{0.13} & N/A \\
    \midrule
    \multicolumn{11}{l}{Neural Scene Representation} \\
    \midrule
    \multirow{3}{*}{\rotatebox{90}{GS}}
    & GS-RelocNet (PR)~\cite{wang2025GSRelocNet}
    & 6min & \checkmark & $\sim$80
    & N/A & \textbf{12}/0.18 & \textbf{13/0.25} & 5/0.19 & 7/0.20 & N/A \\
    & \textbf{PlugGS-Loc~(Ours)}
    & 7min & & 16
    & 11.7/0.08 & 15.9/0.18 & 16.7/0.37 & 2.9/0.13 & 6.0/0.21 & 10.6/0.20 \\
    & \textbf{SplitGS-Loc~(Ours)}
    & 7min & & 97
    & \textbf{9.3/0.06} & 14.1/\textbf{0.17} & 13.8/0.29 & \textbf{2.5/0.12} & \textbf{3.8/0.13} & \textbf{8.7/0.15} \\
    \bottomrule
\end{tabular}
}
\label{tab:main_cambridge}
\end{table}
\begin{table}[t]
\caption{\textbf{Localization results on 7Scenes~\cite{shotton20137scenes}.}
We report median translation and rotation errors.
For all GS-based methods, mapping time includes RGB 3DGS~\cite{kerbl20233dgs} training time (about 6min),
and map size includes per-scene optimized networks and feature fields on top of 3DGS ($\sim$175MB).
$^+$ denotes the version with iterative camera pose refinement using GS-CPR~\cite{liu2025gscpr}, and $^\dagger$ denotes methods that need additional training to adapt the representation for localization per scene.}
\centering
\resizebox{\textwidth}{!}{
\begin{tabular}{clcccccccccccc}
    \toprule
    \multicolumn{1}{l}{} 
    & 
    & \makecell{Mapping\\Time}
    & \makecell{Pose\\Refine}
    & \makecell{Size\\(MB)}
    & \makecell{Chess}
    & \makecell{Fire}
    & \makecell{Heads}
    & \makecell{Office}
    & \makecell{Pumpkin}
    & \makecell{Kitchen}
    & \makecell{Stairs}
    & \makecell{Average\\(cm/$^\circ$)}
    \\ 
    \midrule
    \multirow{3}{*}{\rotatebox{90}{FM}}
    & AS (SIFT)~\cite{sattler2016AS}
    & 0 & & $\sim$200
    & 3/0.87 & 2/1.01 & 1/0.82 & 4/1.15 & 7/1.69 & 5/1.72 & 4/1.01 & 3.7/1.18 \\
    & HLoc (SP+SG)~\cite{sarlin2019hloc, sarlin2020superglue}
    & 0 & & $\sim$2000
    & 2.4/0.84 & 2.3/0.91 & 1.1/0.77 & 3.1/0.92 & 4.9/1.30 & 4.2/1.39 & 5.1/1.41 & 3.3/1.08 \\
    & PixLoc~\cite{sarlin2021pixloc}
    & 0 & \checkmark & $\sim$1000
    & 2/0.80 & 2/0.73 & 1/0.82 & 3/0.82 & 4/1.21 & 3/1.20 & 5/1.30 & 2.9/0.98 \\
    \midrule
    \multirow{2}{*}{\rotatebox{90}{SCR}}
    & DSAC*~\cite{brachmann2021dsacstar}
    & 15h & & 28
    & 0.5/0.17 & 0.8/0.29 & 0.5/0.34 & 1.2/0.35 & 1.2/0.29 & 0.7/0.21 & 2.7/0.78 & 1.1/0.35 \\
    & ACE~\cite{brachmann2023ace}
    & 5min & & 4
    & 0.6/0.18 & 0.8/0.33 & 0.5/0.33 & 1.1/0.29 & 1.1/0.22 & 0.8/0.21 & 2.9/0.81 & 1.1/0.34 \\
    \midrule
    \midrule
    \multicolumn{13}{l}{Neural Scene Representation$^\dagger$} \\
    \midrule
    \multirow{3}{*}{\rotatebox{90}{NeRF}}
    & CROSSFIRE~\cite{moreau2023crossfire}
    & 5h & \checkmark & 50
    & 1/0.4 & 5/1.9 & 3/2.3 & 5/1.6 & 3/0.8 & 2/0.8 & 12/1.9 & 4.4/1.38 \\
    & PNeRFLoc~\cite{zhao2024pnerfloc}
    & hrs & \checkmark & $\sim$50
    & 2/0.80 & 2/0.88 & 1/0.83 & 3/1.05 & 6/1.51 & 5/1.54 & 32/5.73 & 7.3/1.76 \\
    & NeRFMatch~\cite{zhou2024nerfmatch}
    & hrs & \checkmark & $\sim$55
    & 1.0/0.30 & 1.1/0.41 & 1.3/0.92 & 3.1/0.87 & 2.2/0.60 & 1.0/0.28 & 9.3/1.74 & 2.7/0.73 \\
    \midrule
    \multirow{8}{*}{\rotatebox{90}{GS}}
    & GSplatLoc~\cite{sidorov2025gsplatloc}
    & hrs & \checkmark & $>$200
    & 0.4/0.16 & 1.0/0.32 & 1.1/0.62 & 1.9/0.4 & 1.8/0.35 & 2.7/0.55 & 8.8/2.34 & 2.5/0.68 \\
    & GSFFs-PR~\cite{pietrantoni2025gsffs}
    & hrs & \checkmark & $>$200
    & 0.4/0.19 & 0.6/0.26 & 0.5/0.36 & 1.0/0.31 & 1.3/0.38 & 0.6/0.23 & 25/0.63 & 4.2/0.34 \\
    & STDLoc~\cite{huang2025stdloc}
    & 2h & & 16
    & 1.0/0.21 & 0.9/0.30 & 0.6/0.30 & 1.4/0.30 & 1.2/0.22 & 1.4/0.23 & 2.7/0.71 & 1.3/0.32 \\
    & STDLoc~(PR)~\cite{huang2025stdloc}
    & 2h & \checkmark & 888
    & 0.5/0.15 & 0.6/0.24 & 0.5/\textbf{0.26} & 0.9/0.24 & 0.9/0.21 & 0.6/0.19 & 1.4/0.41 & 0.8/0.24 \\
    & GSVisLoc$^+$~\cite{liu2025gscpr, khatib2025gsvisloc}
    & hrs & \checkmark & $>$200
    & 0.4/0.13 & 0.6/0.24 & 0.5/0.34 & 1.0/0.26 & 0.9/0.21 & 0.7/0.18 & 4.7/0.96 & 1.3/0.33 \\
    & RAP$^+$~\cite{liu2025gscpr, Li2025rap}
    & hrs & \checkmark & $\sim$175
    & \textbf{0.3/0.11} & \textbf{0.5/0.21} & \textbf{0.4}/0.27 & \textbf{0.6/0.16} & \textbf{0.8/0.20} & \textbf{0.5/0.12} & \textbf{1.1/0.32} & \textbf{0.6/0.20} \\
    & GS-RelocNet~\cite{wang2025GSRelocNet}
    & hrs & & $>$200
    & 0.4/0.17 & 0.6/0.24 & \textbf{0.4}/0.30 & 0.9/0.24 & 1.0/0.28 & 0.6/0.22 & 1.4/0.39 & 0.8/0.26 \\
    & GS-RelocNet~(PR)~\cite{wang2025GSRelocNet}
    & hrs & \checkmark & $>$200
    & 0.4/0.15 & 0.6/\textbf{0.21} & \textbf{0.4/0.26} & 0.9/0.24 & 0.9/0.21 & 0.6/0.18 & 1.3/0.35 & 0.7/0.23 \\
    \midrule
    \multicolumn{13}{l}{Neural Scene Representation} \\
    \midrule
    \multirow{5}{*}{\rotatebox{90}{GS}}
    & GS-RelocNet~\cite{wang2025GSRelocNet}
    & 6min & & $\sim$175
    & 1.3/0.81 & 1.2/0.65 & 0.7/0.73 & 1.6/0.89 & 2.7/1.01 & 2.5/1.10 & 2.1/1.00 & 1.7/0.88 \\
    & GS-RelocNet~(PR)~\cite{wang2025GSRelocNet}
    & 6min & \checkmark & $\sim$175
    & 1.0/0.72 & 1.0/0.64 & 0.6/0.70 & 1.4/0.82 & 2.2/0.93 & 2.0/1.02 & 1.9/0.92 & 1.4/0.82 \\
    & \textbf{PlugGS-Loc~(Ours)}
    & 7min & & 16
    & 0.9/0.18 & 0.9/0.31 & 0.7/0.30 & 1.4/0.30 & 1.6/0.27 & 1.3/0.23 & 1.8/0.50 & 1.2/0.30 \\
    & \textbf{SplitGS-Loc~(Ours)}
    & 7min & & 88
    & \textbf{0.7/0.16} & \textbf{0.7/0.29} & \textbf{0.6/0.26} & \textbf{1.2/0.25} & \textbf{1.4/0.25} & \textbf{1.0/0.18} & \textbf{1.6/0.47} & \textbf{1.0/0.26} \\
    \bottomrule
\end{tabular}
}
\label{tab:main_7scenes}
\end{table}
\subsection{Experimental setup}
\label{sec:setup}
\paragraph{Datasets.}
We conducted experiments on Cambridge Landmarks~\cite{kendall2015cambridge} and 7Scenes \cite{shotton20137scenes} datasets.
Cambridge Landmarks consists of five outdoor scenes, with scene scales ranging from $875\text{m}^2$ to $7600\text{m}^2$.
In contrast, 7Scenes includes seven indoor scenes whose scales range from $1\text{m}^2$ to $18\text{m}^2$.
For the ground-truth camera poses, we adopted the SfM-based annotations reported by~\cite{Brachmann2021pseudoGT}, as the D-SLAM-based ground truths are unreliable for neural rendering~\cite{liu2025gscpr}.

\paragraph{Implementation details.}
We basically follow the setting of STDLoc~\cite{huang2025stdloc}.
For the weight aggregation, the weight threshold $\tau$ in Eq.~\ref{eq:primary_gaussians} is fixed at 0.1 across all scenes to avoid per-scene optimization.
We pre-filter superfluous Gaussians via proposed weight aggregation process before PlugGS-Loc and SplitGS-Loc to obtain a clean weight set.
In SplitGS-Loc, the number of anchor Gaussians retained during sampling is doubled to ensure sufficient spatial coverage.
The splitting factor $\beta$ in Eq.~\ref{eq:child_definition} is set to 1.4 to extend the side components further along the major axis while preserving the approximation quality under the positive semi-definiteness constraint.
For feature aggregation and inference, we use the same image feature extractor and PnP+RANSAC algorithm as STDLoc, SuperPoint~\cite{detone2018superpoint} and PoseLib~\cite{PoseLib}, respectively.
All experiments are conducted on a single NVIDIA GeForce RTX 4090.

\paragraph{Metrics.}
Our GSFFs construction on top of a pre-trained 3DGS completes within a minute,
where the reported mapping time additionally includes RGB 3DGS~\cite{kerbl20233dgs} training time (about 6min) shared across all GS-based methods for fair comparison.
Mapping time and map sizes are estimated according to the data required at inference. 
Details are provided in the appendix.

\subsection{Comparative results}
\label{sec:results}
Our proposed frameworks show superior performance on both indoor and outdoor scenes as shown in Tab.~\ref{tab:main_7scenes} and Tab.~\ref{tab:main_cambridge}.
Notably, SplitGS-Loc achieves remarkable performance on the largest scene, GreatCourt, without per-scene training for feature fields or pose refinement.
These results demonstrate that accurate localization can be achieved without relying on heavy GSFFs or iterative refinement, 
moving toward a streamlined solution for real-world visual localization: smaller, faster, yet more precise.

\subsection{Ablation studies}
\label{sec:ablation}
\begin{figure}[t]
\centering
\includegraphics[width=1.0\linewidth]{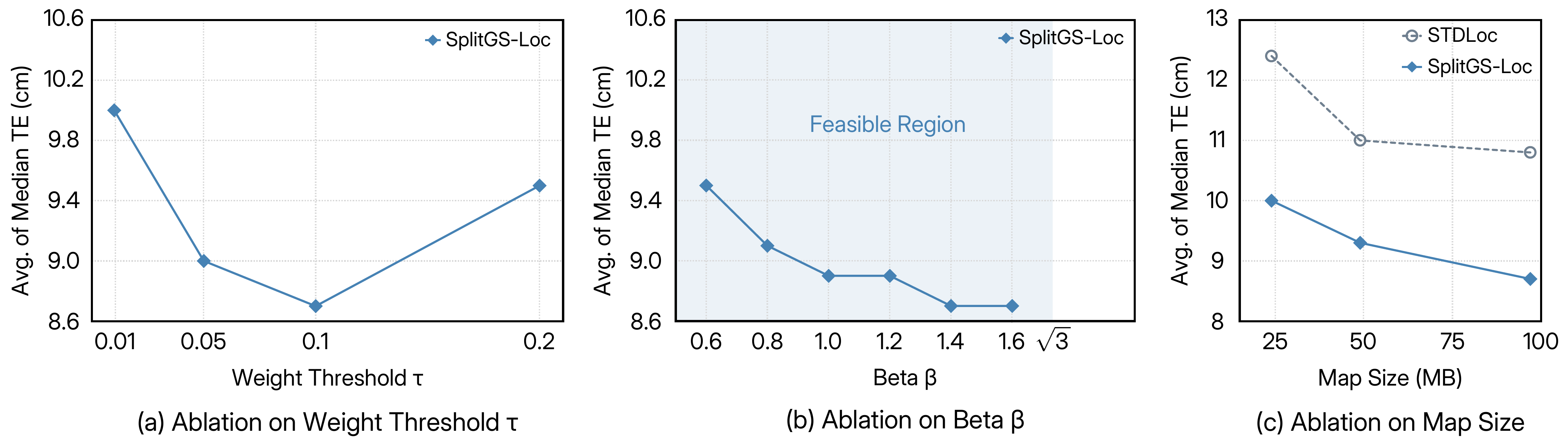}
\caption{\textbf{Ablation on mapping hyperparameters.}
We compare the average per-scene median translation error under varying weight thresholds $\tau$, splitting factor $\beta$, and the map size on Cambridge Landmarks~\cite{kendall2015cambridge}.
(a) Mild pruning causes feature blurring and aggressive pruning limits coverage of the scene, bringing suboptimal results.
(b) Localization accuracy increases monotonically with $\beta$ within the feasible region, suggesting that wider separation of child Gaussians more effectively alleviates pixel-Gaussian matching ambiguity.
(c) SplitGS-Loc scales consistently beyond the saturation by effective densification at the level of detail.
}
\label{fig:ablation}
\end{figure}

\paragraph{Weight threshold $\tau$.}
We conduct an ablation on the weight threshold $\tau$ that determines the importance of each Gaussians, testing values of ${0.01, 0.05, 0.1, 0.2}$, as depicted in Fig.~\ref{fig:ablation} (a).
When the pruning is too mild, features that have weak pixel-Gaussians associations are aggregated together, resulting in blurred representations and thus degraded localization accuracy.
This reflects a limitation inherent to rendering-based GSFFs optimization as discussed in Sec.~\ref{sec:preliminary_gsff}, where volumetric rendering disperses gradients across numerous Gaussians per ray and weakens feature discriminativeness.
Conversely, when the pruning is too aggressive, the absolute number of remaining Gaussians and their spatial coverage become insufficient for stable pose estimation.
Meanwhile, our methods are robust to this hyperparameter within a reasonable range, with $0.05$ performing on par with $0.1$.

\paragraph{Splitting factor $\beta$.}
Fig.~\ref{fig:ablation} (b) illustrates the effect of the splitting factor $\beta$ that controls the side offsets of child Gaussians.
Localization accuracy improves monotonically as $\beta$ increases within the feasible region ($\beta \in (0, \sqrt{3})$), where the positive semi-definiteness constraint is satisfied.
This suggests that, once the moment matching approximation holds, placing child Gaussians further apart along the major axis more effectively reduces pixel-Gaussian matching ambiguity.

\paragraph{Component ablation.}
Tab.~\ref{tab:alternative} (a) provides a separate ablation using STDLoc~\cite{huang2025stdloc} as the baseline, independently swapping the GSFFs and the sampling strategy.
Our GSFFs yields a significant accuracy improvement from 13.1cm/0.22$^\circ$ to 9.6cm/0.17$^\circ$, demonstrating that the proposed framework is not only efficient but also more effective.
In parallel, our weight-based sampling strategy brings additional consistent improvement, achieving the best performance of 8.7cm/0.15$^\circ$.
The result confirms that the composition weight acts as a unified criterion that simultaneously guides feature registration and Gaussian sampling, collectively enabling a more accurate and efficient GSFFs construction.

\subsection{Analysis}
\label{sec:analysis}
\begin{table}[t]
  \caption{\textbf{Alternative methods.}
  We compare alternative designs for SplitGS-Loc on Cambridge Landmarks~\cite{kendall2015cambridge}.
  (a) Our GSFFs and sampling strategy bring complementary improvements without multi-stage training.
  (b) Replacing Gaussian attributes with position-only alternatives degrades accuracy.}
  \label{tab:alternative}
  \centering
  \small
  \begin{minipage}[t]{0.44\linewidth}
    \centering
    \textbf{(a)} Ablation with STDLoc~\cite{huang2025stdloc}.\\
    \resizebox{\linewidth}{!}{
    \begin{tabular}{ccc}
      \toprule
      GSFFs & Sampling & Average (cm/$^\circ$) \\
      \midrule
      STDLoc & Cosine Similarity & 13.1/0.22 \\
      STDLoc & Eq.~\ref{eq:gussian_score} & 12.8/0.22 \\
      Ours & Cosine Similarity & 9.6/0.17 \\
      Ours & Eq.~\ref{eq:gaussian_score_split} & \textbf{8.7/0.15} \\
      \bottomrule
    \end{tabular}}
  \end{minipage}
  \hfill
  \begin{minipage}[t]{0.55\linewidth}
    \centering
    \textbf{(b)} Ablation on Gaussian Attributes. \\
    \resizebox{\linewidth}{!}{
    \begin{tabular}{ccc}
      \toprule
      Feature Aggregation & Densification & Average (cm/$^\circ$) \\
      \midrule
      Projection+Average & - & 14.1/0.26 \\
      PlugGS-Loc (Ours) & - & 10.6/0.20 \\
      Ours & NN Upsampling & 10.0/0.18 \\
      SplitGS-Loc (Ours) & Eq.~\ref{eq:split_definition}, Eq.~\ref{eq:child_definition} & \textbf{8.7/0.15} \\
      \bottomrule
    \end{tabular}}
  \end{minipage}
\end{table}
\paragraph{Map capacity.}
Larger maps generally lead to better performance, similar to how dense reconstruction-based pipelines outperform sparse ones.
Accordingly, the key questions are (i) how they perform under the same map capacity and (ii) how effectively the performance scales with increasing map size.
To examine this, as illustrated in Fig.~\ref{fig:ablation} (c), we compare direct 2D-3D matching pipelines using similar map sizes between STDLoc~\cite{huang2025stdloc} and proposed SplitGS-Loc.
For fairness, we exclude any scene-specific post-training and adjust the number of sampled Gaussians to approximate the same map size, using $3\times2^{13}$, $3\times2^{14}$, $3\times2^{15}$ samples for STDLoc, and $2^{13}, 2^{14}, 2^{15}$ (one third) for SplitGS-Loc as each Gaussian is further divided into three child Gaussians.
Fig.~\ref{fig:ablation} (c) shows that merely increasing the number of sampled Gaussians faces an inherent limitation: the performance saturates.
By contrast, our SplitGS-Loc exhibits sustained improvements as the map size increases, highlighting the efficacy of our splitting strategy.

\paragraph{Advantages of Gaussian attributes.}
We investigate the advantages of leveraging Gaussian attributes for localization.
Tab.~\ref{tab:alternative} (b) presents results when Gaussian attributes are replaced with geometry-only alternatives under the same pipeline.
First, we replace our rasterization-based pipeline with Projection+Average.
Our method filters strong pixel-Gaussian correspondences via composition weights and aggregates features through soft assignment, whereas Projection+Average directly projects Gaussian centers onto training views and uniformly averages the nearest keypoint descriptors.
This leads to a significant median error increase from 10.6cm/0.20$^\circ$ to 14.1cm/0.26$^\circ$; composition weights derived from opacity and transmittance naturally account for occlusion and visibility, whereas geometric projection alone cannot.

Besides, replacing our Gaussian Splitting with k-nearest neighbor upsampling ($k=3$) results in 10.0cm/0.18$^\circ$, which inserts new Gaussians at midpoint positions between each Gaussian and its nearest neighbors with linearly interpolated features, compared to 8.7cm/0.15$^\circ$ for SplitGS-Loc.
This suggests that proposed Mixture-of-Gaussians-based splitting guided by scale enables more accurate densification.
In summary, exploiting composition weights and volumetric attributes during mapping yields substantially more discriminative feature fields.

\begin{table}[t]
\caption{\textbf{PnP convergence.}
We investigate the correlation between PnP convergence and accuracy on Cambridge Landmarks~\cite{kendall2015cambridge},
comparing STDLoc~\cite{huang2025stdloc} and our SplitGS-Loc under efficient (a,c) and default (b,d) setting.
By mitigating the pixel-Gaussian matching ambiguity, 
SplitGS-Loc produces fewer many-to-one inputs and more valid inliers, achieving the best accuracy with stable convergence.}
\centering
\resizebox{\textwidth}{!}{
\begin{tabular}{clcccccc}
    \toprule
    \multicolumn{2}{l}{\makecell{Direct 2D-3D\\Matching}} 
    & \makecell{RANSAC\\Min. Iter.} 
    & \makecell{RANSAC\\Max. Iter.}
    & \makecell{\# of PnP\\Many-to-one ($\downarrow$)}
    & \makecell{\# of PnP\\Final Inliers ($\uparrow$)}
    & \makecell{Runtime ($\downarrow$)\\(ms)}
    & \makecell{Average ($\downarrow$)\\(cm / $^\circ$)}
    \\ 
    \midrule
    (a) 
    & STDLoc
    & $10^2$
    & $10^3$
    & 313
    & 362
    & 54.4
    & 12.8/0.21
    \\
    (b)
    & STDLoc
    & $10^3$
    & $10^5$
    & 313
    & 367
    & 314.4
    & 11.2/0.20
    \\
    \midrule
    (c)
    & SplitGS-Loc (Ours)
    & $10^2$
    & $10^3$
    & 199
    & 447
    & 54.8
    & 8.9/\textbf{0.15}
    \\
    (d)
    & SplitGS-Loc (Ours)
    & $10^3$
    & $10^5$
    & 199
    & 450
    & 215.5
    & \textbf{8.7/0.15}
    \\
    \bottomrule
\end{tabular}
}
\label{tab:pnp_convergence}
\end{table}
\paragraph{PnP convergence.}
Tab.~\ref{tab:pnp_convergence} presents a PnP convergence analysis of direct 2D-3D matching-based methods under different RANSAC settings.
While STDLoc (a, b) struggles to reach consensus due to redundant many-to-one matches,
our SplitGS-Loc (c, d) mitigates the pixel-Gaussian matching ambiguity, reducing many-to-one inputs from 313 to 199.
This contributes to obtaining more inliers, better accuracy, and faster convergence: from 314ms to 215ms under the same setting.
This stable convergence in turn yields robustness under an efficient setting:
while STDLoc suffers a notable accuracy drop from 11.2cm to 12.8 cm (b vs a),
SplitGS-Loc remains stable about 8.7cm (d vs c), suggesting that its robustness stems from overall high-quality matches.

\paragraph{Limitation.}
\label{sec:limitation}

While the proposed framework refines photometrically optimized Gaussians, it is not entirely independent of the quality of original Gaussians.
As such, scenes with severe illumination changes or sparse trajectories over large-scale environments lie beyond what can be meaningfully examined with current GS techniques, as GS struggles to converge in such settings.
This dependency is shared across neural scene representation-based localization, which has so far been validated only on small- to medium-scale scenes.
Still, our conversion takes under a minute and applies to any explicit Gaussian formulation, so it can be readily plugged in whenever a well-trained large-scale 3DGS becomes available.
Given the rapid progress of neural scene representation research, extending to such challenging real-world conditions would be an interesting direction for future work.

\section{Conclusion}
\label{sec:conclusion}

Recent years have seen significant progress in Gaussian Splatting-based representations, and their potential for visual localization continues to broaden.
Yet photometrically optimized GSFFs are fundamentally ill-suited for 2D-3D matching, as redundant Gaussians and many-to-one pixel-Gaussian correspondences undermine both matching robustness and PnP convergence.
This work has proposed SplitGS-Loc, which resolves these limitations effectively by exploiting Gaussian attributes for localization-specialized GSFFs construction, eliminating the need for per-scene multi-stage optimization.
By disambiguating pixel-Gaussian correspondences, it ensures stable PnP convergence and sustained accuracy gains as the map scales, achieving state-of-the-art performance at a fraction of the cost that prior pipelines require.
Beyond localization accuracy, this work reinforces long-standing goals in visual localization: efficiency and practicality.

\section*{Acknowledgement}

This work was supported in part by MSIT/IITP (No. RS-2022-II220680, RS-2020-II201821, RS-2019-II190421, RS-2024-00459618, RS-2024-00360227, RS-2024-00437633, RS-2026-25549037, RS-2025-25442569), MSIT/NRF (No. RS-2024-00357729), and KNPA/KIPoT (No. RS-2025-25393280).

{
    \small
    \bibliographystyle{unsrtnat}
    \bibliography{main}
}

\appendix
\clearpage
\setcounter{page}{1}
\setcounter{section}{0}
\renewcommand{\thesection}{\Alph{section}}
\setcounter{figure}{0}
\renewcommand{\thefigure}{A\arabic{figure}}
\setcounter{table}{0}
\renewcommand{\thetable}{A\arabic{table}}

\section{Derivation of the Splitting Parameters}
\label{sec:moment_matching}

This section provides a detailed derivation of the mixture parameters obtained by matching the first four moments.
As mentioned in the main paper, we define a parent 1D Gaussian distribution in the canonical space as $\tilde{p}(x) \sim \mathcal{N}(0,s^2)$,
and approximate it by a symmetric three-component mixture $\tilde{q}(x)$ as:
\begin{equation*}
    \tilde{p}(x) \approx \tilde{q}(x) = \lambda q_{-}(x) + \lambda_\text{o} q_\text{o}(x) + \lambda q_{+}(x),
\end{equation*}
\begin{equation*}
    \begin{aligned}
    q_{-}(x) &\sim \mathcal{N}(-\beta s,\ \sigma^2),\\
    q_\text{o}(x) &\sim \mathcal{N}(0,\ \sigma_\text{o}^2),\\
    q_{+}(x) &\sim \mathcal{N}(+\beta s,\ \sigma^2).
    \end{aligned}
\end{equation*}
Here $\lambda,\lambda_\text{o} \in (0,1)$ are mixture weights satisfying $2\lambda + \lambda_\text{o} = 1$, and $\beta>0$ controls the side offsets.

\paragraph{Balanced Splitting.}
The purpose of splitting is to densify the 3DGS such that, after rendering, pixel-Gaussian associations become more localized and less ambiguous.
From this perspective, an important requirement for the approximation $\tilde q(x)$ is that the three components $q_-(x)$, $q_\text{o}(x)$, and $q_+(x)$ share the spatial extent of the original density in a balanced manner.
If the central component $q_\text{o}(x)$ were given a substantially larger variance than the side components, most of the probability mass would remain concentrated around the original mean.
Then, the side components would contribute negligibly and tend to be pruned by the subsequent weight aggregation process.
Conversely, if the side components $q_{-}(x)$ and $q_{+}(x)$ were overly spread, the mixture would deviate from the original Gaussian and degrade the approximation quality.
To avoid these issues, we first impose $\sigma_\text{o} = \sigma$, thereby reducing the degrees of freedom and biasing the solution toward our final goal, fine-grained localization.

\paragraph{Moment Matching.}
Since both distributions are symmetric around the origin, all odd-order moments are zero.
In the following, we therefore describe how to match the second and fourth moments.
For a univariate Gaussian random variable $Z \sim \mathcal{N}(m, v)$, the second and fourth moments are
\begin{equation}
    \mathbb{E}[Z^2] = v + m^2, \ \ 
    \mathbb{E}[Z^4] = 3v^2 + 6vm^2 + m^4.
\end{equation}
To match the second-order moments of $\tilde{p}(x)$ and $\tilde{q}(x)$, the following constraint is imposed:
\begin{equation}
    \mathbb{E}_{\tilde{p}}[x^2] = \mathbb{E}_{\tilde{q}}[x^2].
    \label{eq:second_moment_matching}
\end{equation}
Since $\tilde{p}(x) \sim \mathcal{N}(0,s^2)$, we have
\begin{equation}
    \mathbb{E}_{\tilde{p}}[x^2] = s^2.
    \label{eq:second_moment_raw_p}
\end{equation}
For the mixture $\tilde{q}(x)$, the second moment can be written as
\begin{equation}
    \begin{aligned}
        \mathbb{E}_{\tilde{q}}[x^2]
        &= \lambda\,\mathbb{E}_{q_{-}}[x^2]
          + \lambda_\text{o}\,\mathbb{E}_{q_\text{o}}[x^2]
          + \lambda\,\mathbb{E}_{q_{+}}[x^2] \\
        &= 2\lambda(\sigma^2 + \beta^2 s^2) + \lambda_\text{o} \sigma^2.
    \end{aligned}
    \label{eq:second_moment_raw_q}
\end{equation}
Substituting Eq.~\ref{eq:second_moment_raw_p} and Eq.~\ref{eq:second_moment_raw_q} into Eq.~\ref{eq:second_moment_matching} yields
\begin{equation}
    s^2 = 2\lambda(\sigma^2 + \beta^2 s^2) + \lambda_\text{o} \sigma^2.
    \label{eq:second_moment_raw}
\end{equation}
In parallel, the fourth-order moment matching between $\tilde{p}(x)$ and $\tilde{q}(x)$ is enforced by requiring
\begin{equation}
    \mathbb{E}_{\tilde{p}}[x^4] = \mathbb{E}_{\tilde{q}}[x^4].
    \label{eq:fourth_moment_matching}
\end{equation}
The fourth moment of $\tilde{p}(x)$ is
\begin{equation}
    \mathbb{E}_{\tilde{p}}[x^4] = 3s^4, 
    \label{eq:fourth_moment_raw_p}
\end{equation}
and for $\tilde{q}(x)$, we have
{\small
\begin{equation}
    \begin{aligned}
    \mathbb{E}_{\tilde{q}}[x^4]
    &= \lambda\, \mathbb{E}_{q_{-}}[x^4]
      + \lambda_\text{o}\, \mathbb{E}_{q_\text{o}}[x^4]
      + \lambda\, \mathbb{E}_{q_{+}}[x^4] \\
    &= 2\lambda \bigl(3\sigma^4 + 6\sigma^2(\beta s)^2 + (\beta s)^4 \bigr)
      + \lambda_\text{o} \bigl(3\sigma^4\bigr).
    \label{eq:fourth_moment_raw_q}
    \end{aligned}
\end{equation}
}
From Eq.~\ref{eq:fourth_moment_raw_p} and Eq.~\ref{eq:fourth_moment_raw_q} together with Eq.~\ref{eq:fourth_moment_matching},
\begin{equation}
    3s^4
    = 2\lambda \bigl(3\sigma^4 + 6\sigma^2(\beta s)^2 + (\beta s)^4 \bigr)
      + \lambda_\text{o} \bigl(3\sigma^4\bigr).
    \label{eq:fourth_moment_raw}
\end{equation}
For simplicity, we introduce the dimensionless variable $U = \sigma^2/s^2$.
Using $2\lambda + \lambda_\text{o} = 1$, the second-order constraint in Eq.~\ref{eq:second_moment_raw} then becomes
\begin{equation}
    (1 - \lambda_\text{o})(U + \beta^2) + \lambda_\text{o} U = 1,
    \label{eq:second_moment_dimless_U}
\end{equation}
and the fourth-order constraint in Eq.~\ref{eq:fourth_moment_raw} becomes
\begin{equation}
    (1 - \lambda_\text{o}) \bigl(U^2 + 2U\beta^2 + \frac{\beta^4}{3}\bigr)
    + \lambda_\text{o} U^2 = 1.
    \label{eq:fourth_moment_dimless_U}
\end{equation}
Solving Eq.~\ref{eq:second_moment_dimless_U} for $U$ gives
\begin{equation}
    \begin{aligned}
        U = 1 - (1 - \lambda_\text{o}) \beta^2.
    \end{aligned}
    \label{eq:U_from_second}
\end{equation}
We then expand the left-hand side of Eq.~\ref{eq:fourth_moment_dimless_U} as
\begin{equation}
        (1-\lambda_\text{o})\bigl( (U+\beta^2)^2 - \frac{2}{3}\beta^4 \bigr) + (\lambda_\text{o}U^2 - 1) = 0,
        \label{eq:fourth_moment_dimless_U_left}
\end{equation}
and substitute Eq.~\ref{eq:U_from_second} into the left and right terms of Eq.~\ref{eq:fourth_moment_dimless_U_left} as
\begin{equation}
    \begin{aligned}
        &(1-\lambda_\text{o})\bigl( (U+\beta^2)^2 - \frac{2}{3}\beta^4 \bigr) \\
        = &(1-\lambda_\text{o})(1 + 2\lambda_\text{o} \beta^2 + \lambda_\text{o}^2 \beta^4 - \frac{2}{3}\beta^4),
    \end{aligned}
    \label{eq:fourth_moment_left_term}
\end{equation}
and
\begin{equation}
    \begin{aligned}
        &\lambda_\text{o}U^2 - 1 \\
        = &\lambda_\text{o}\bigl( 1 - 2(1 - \lambda_\text{o})\beta^2 + (1-\lambda_\text{o})^2\beta^4 \bigr) - 1 \\
        = &(1 - \lambda_\text{o}) ( -1-2\lambda_\text{o} \beta^2 + \lambda_\text{o} \beta^4 - \lambda_\text{o}^2 \beta^4 ),
    \end{aligned}
    \label{eq:fourth_moment_right_term}
\end{equation}
respectively.
Combining Eq.~\ref{eq:fourth_moment_left_term} and Eq.~\ref{eq:fourth_moment_right_term} yields
\begin{equation}
    (1-\lambda_\text{o})(-\frac{2}{3}\beta^4 + \lambda_\text{o}\beta^4) = 0,
    \label{eq:fourth_moment_reduced}
\end{equation}
and finally we obtain a polynomial in $\lambda_\text{o}$,
\begin{equation}
    \begin{aligned}
        \frac{\beta^4}{3}\,(3\lambda_\text{o} - 2)(\lambda_\text{o} - 1) = 0.
    \end{aligned}
\end{equation}
The $\lambda_\text{o} = 1$ corresponds to the degenerate case $\lambda = 0$, where the side components vanish and no splitting occurs.
We therefore take $\lambda_\text{o} = 2/3$, and $U = \sigma^2 / s^2 = 1 - \beta^2/3$.
The covariance matrix is required to be positive semi-definite, which in our setting is equivalent to
$\sigma^2 > 0$, \ie, $1 - \beta^2/3 > 0$ and hence $\beta \in (0, \sqrt{3})$.
Under this feasible range, we set $\beta=1.4$, which places the side components further apart along the major axis while satisfying the positive semi-definiteness constraint, as validated by the ablation study in the main paper.
\section{Mapping time}
\label{sec:map}

\paragraph{Per-stage Processing Time.}
\begin{table}[b]
\caption{\textbf{Per-stage processing time.}}
\centering
\begin{tabular}{@{\hspace{10pt}}l@{\hspace{40pt}}c@{\hspace{10pt}}}
    \toprule
    Stage
    & Time (s)
    \\ 
    \midrule
    Gaussian Splitting
    & 0.4
    \\
    Pre-filtering
    & 10.6
    \\
    Weight Aggregation
    & 9.8
    \\
    Gaussian Sampling
    & 10.2
    \\
    Feature Registration
    & 33.4
    \\
    \midrule
    Total
    & 64.4
    \\
    \bottomrule
\end{tabular}
\label{tab:stage_mapping_time}
\end{table}
Tab.~\ref{tab:stage_mapping_time} reports the processing time of each stage in SplitGS-Loc, including Gaussian splitting, pre-filtering, weight aggregation, Gaussian sampling, and feature registration. 
Each stage is averaged across the Cambridge Landmarks~\cite{kendall2015cambridge}. 
If image features are precomputed, the feature registration cost can be further reduced; however, we do not account for this optimization, assuming only raw datasets are provided.

\paragraph{Mapping Time.}
We set a SfM output as a zero across all methods and measure additional per-scene training (or post-processing) time.
Accordingly, for all GS-based methods, the mapping time includes the time required to train the RGB 3DGS~\cite{kerbl20233dgs} (approximately 6 minutes per scene), even though it is not trained exclusively for localization.
Although we make considerable effort to ensure a fair comparison across methods, achieving a fully fair comparison of mapping times is difficult in practice, since the mapping times of each method are highly sensitive to hardware, implementation details, and training schedules.
For this reason, when a paper explicitly reports mapping times, we prioritize those published numbers~\cite{brachmann2023ace,moreau2023crossfire,huang2025stdloc}.

For methods that employ a scene-specific model for pose initialization before pose refinement, we treat the training time of the initialization module as the per-scene mapping time. 
This is aligned with the protocol adopted in CROSSFIRE~\cite{moreau2023crossfire}, which first introduced the neural scene representation-based relocalization, where the training time of the absolute pose regression model is included in the overall mapping cost. 
This also supports a fair comparison with direct 2D-3D matching pipelines such as STDLoc~\cite{huang2025stdloc} and ours, which do not rely on any pose priors.
For the remaining works, which provide no evidence of training completed in under an hour but involve multiple optimization epochs, we estimate their per-scene mapping process likely requires several hours.

\paragraph{Map size.}
Map sizes for scene coordinate regression and feature matching-based methods are referenced from ACE~\cite{brachmann2023ace} and GLACE~\cite{wang2024glace}, 
while those for NeRF-based methods are standardized following the specifications in CROSSFIRE~\cite{moreau2023crossfire} and NeRFMatch~\cite{zhou2024nerfmatch}.
For 3DGS-based methods, the map size is estimated approximately according to the data required at inference time:
(i) if photometric rendering is required, the full RGB 3DGS is counted,
(ii) if feature rendering is required, the full GSFFs are included, and
(iii) for direct 2D-3D matching-based methods, only the 3D points and corresponding features are considered.
\section{Quantitative Analysis}
\begin{figure*}[t]
\centering
\includegraphics[width=1.0\linewidth]{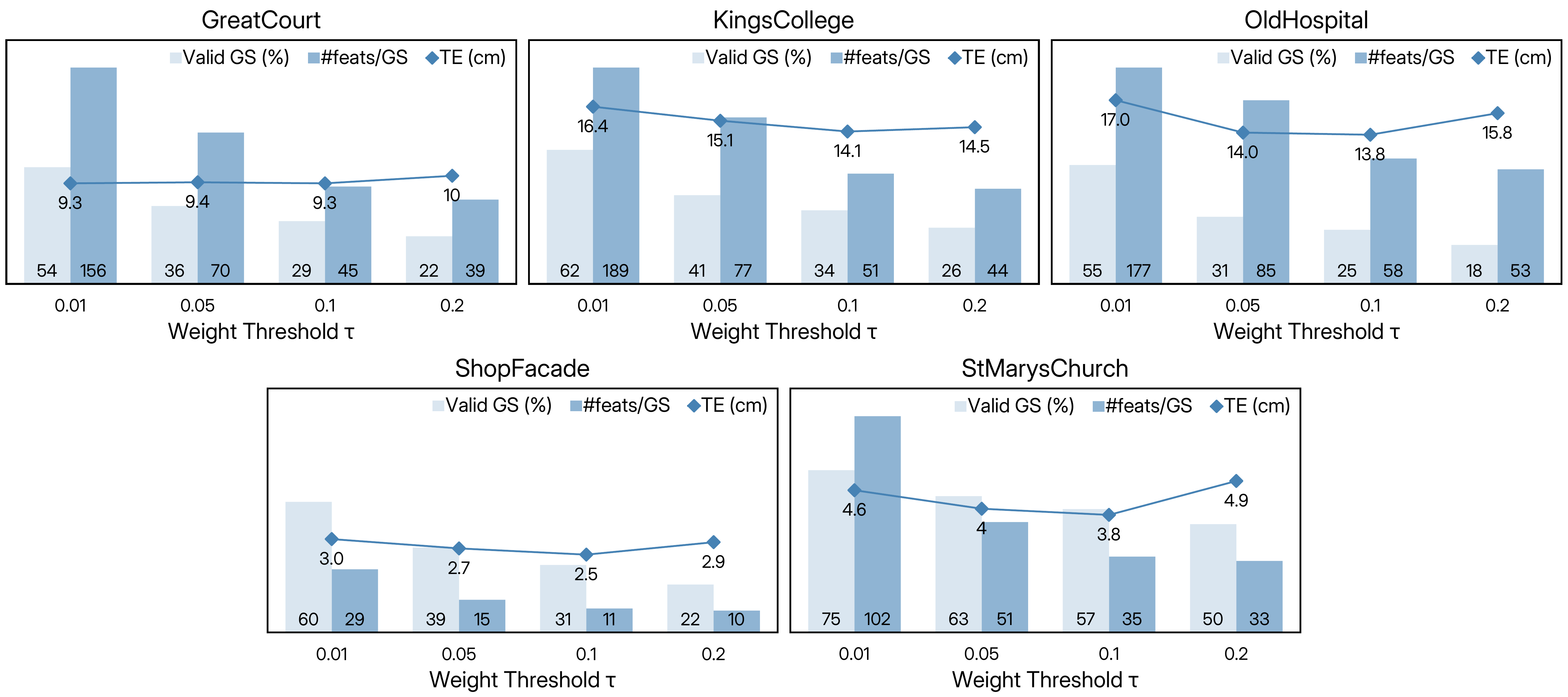}
\caption{
\textbf{Analysis of weight thresholding in SplitGS-Loc.}}
\label{fig:featgs_splitgs-loc}
\end{figure*}

This section extends the ablation on the weight threshold in Sec.~\ref{sec:ablation} by providing a detailed analysis of how the primary Gaussian selection in Eq.~\ref{eq:primary_gaussians} transforms a full photometric GS into localization-specialized GSFFs.
Fig.~\ref{fig:featgs_splitgs-loc} reports three statistics from the weight threshold $\tau$:
(i) the ratio of Gaussians whose informative weight set $\mathcal{W}_n$ (Eq.~\ref{eq:informative_weight_set}) is non-empty (Valid GS),
(ii) the average number of multi-view features aggregated per valid Gaussian, \ie, the average cardinality of the non-empty sets (feats/GS),
and (iii) the median translation errors.

Although the scenes differ in terms of trajectories as well as environmental changes, we observe several trends that remain broadly consistent across scenes.
First, even when preserving Gaussians that contribute at least $1\%$ to any pixel ($\tau = 0.01$), about $40\%$ of the Gaussians are excluded across the scenes. 
This result indicates that the photometric GS contains substantial redundancy from the perspective of solving the matching task, which requires strong correspondences.

Subsequently, we validate that simply aggregating a large number of multi-view features into each Gaussian does not directly guarantee improved localization accuracy in most scenes.
Rather, as discussed in the main paper, maintaining a sufficient valid Gaussian ratio, \ie, adequate spatial coverage of the scene, is critical.
Finally, despite the diversity across scenes, the effective range of $\tau$ values does not vary dramatically.
This suggests that the proposed composition weight-based GSFFs construction framework can be applied to new scenes without per-scene tuning of the weight threshold.

\section{Qualitative Analysis}
\begin{figure*}[t]
\centering
\includegraphics[width=1.0\linewidth]{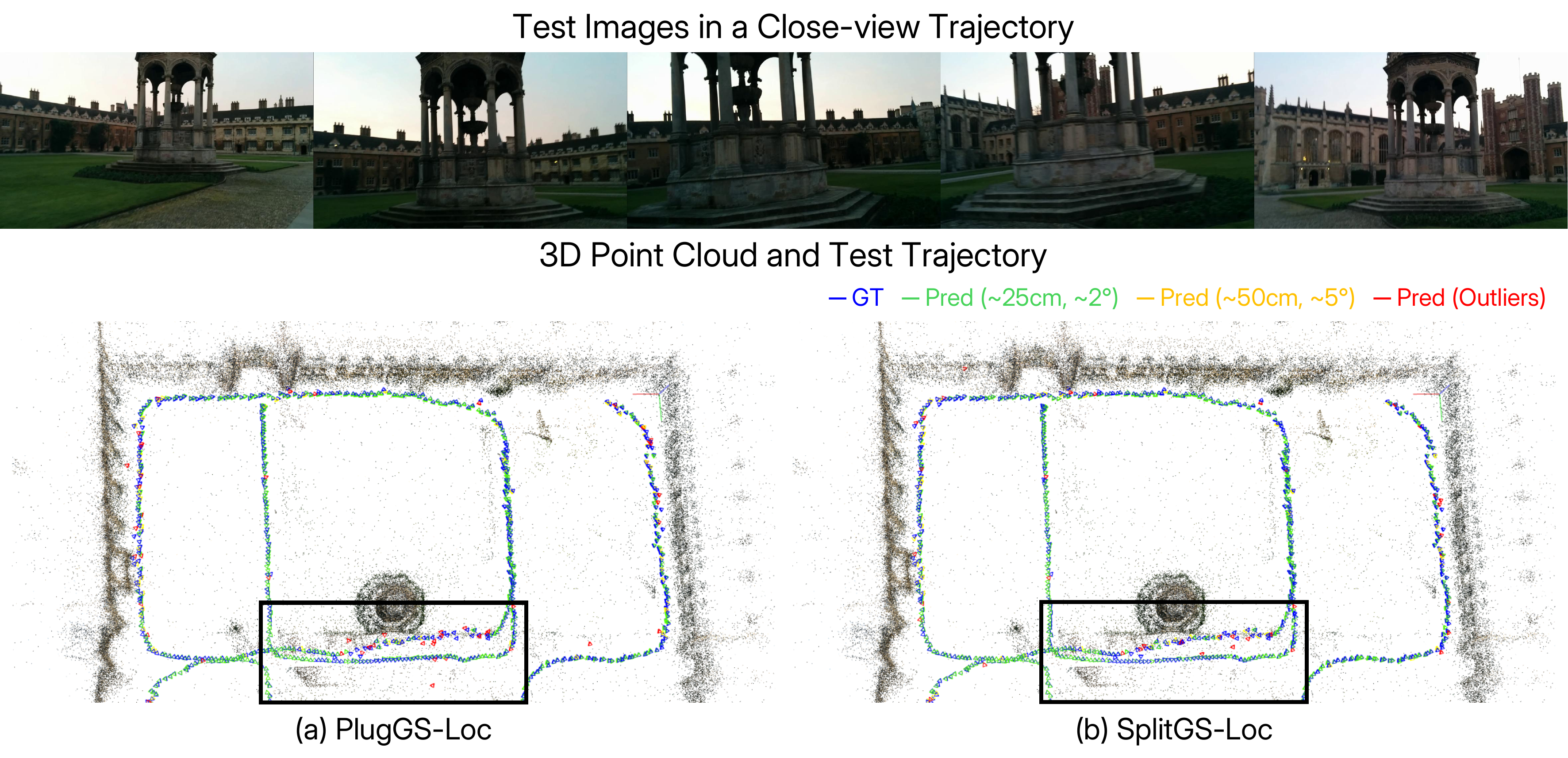}
\caption{
\textbf{Localization results of PlugGS-Loc and SplitGS-Loc on GreatCourt.}}
\label{fig:greatcourt}
\end{figure*}

Fig.~\ref{fig:greatcourt} presents localization results on GreatCourt, the largest and most challenging outdoor scene in the Cambridge Landmarks dataset~\cite{kendall2015cambridge}. 
We color each predicted camera pose according to its translation and rotation errors, following common recall thresholds in large-scale visual localization: green for poses within $(25\text{cm}, 2^\circ)$, yellow for poses within $(50\text{cm}, 5^\circ)$, and red otherwise. 
Without any training for feature fields, PlugGS-Loc produces stable pose predictions along most of the trajectory. 
Nevertheless, it becomes less reliable when the camera observes the foreground at very close range, as shown in Fig.~\ref{fig:greatcourt}. 
This originates from pixel-Gaussian matching ambiguity discussed in Fig.~\ref{fig:motivation}.
When photometrically optimized Gaussians are directly used for localization, those close to the camera tend to cover larger image regions and thus become associated with many pixels, exacerbating many-to-one pixel-Gaussian correspondences and destabilizing PnP convergence.
Adopting SplitGS-Loc as introduced in Sec.~\ref{sec:method_splitgs-loc} stabilizes pose estimation for these challenging views, as illustrated in Fig.~\ref{fig:greatcourt}.
\section{Pose Refinement}
\label{sec:suppl_refinement}

While SplitGS-Loc directly estimates the camera pose through 2D-3D matching using a compact feature map, its output can also serve as an initial pose for existing refinement methods.
To examine the complementarity of SplitGS-Loc with those methods, we apply the feature rendering-based refinement of STDLoc~\cite{huang2025stdloc}, which renders feature and depth maps from the full GSFFs, and the RGB rendering-based refinement of GS-CPR~\cite{liu2025gscpr}, which establishes image correspondences between the rendered RGB image and the query using MASt3R~\cite{leroy2024mast3r}.
As shown in Tab.~\ref{tab:suppl_refinement}, feature rendering-based refinement consistently improves SplitGS-Loc on both datasets.
This improvement, however, requires storing the full GSFFs, drastically increasing the map size.
GS-CPR further improves the 7Scenes result, while degrading performance on Cambridge Landmarks; we suspect that this difference stems from the fixed input resolution of MASt3R relative to the image resolution of each dataset.
Overall, these results suggest that pose refinement can further improve SplitGS-Loc, but the gain comes with additional map storage and depends on the specific refinement method and its limitations.

\begin{table}[t]
\caption{\textbf{Pose refinement results on Cambridge Landmarks~\cite{kendall2015cambridge} and 7Scenes~\cite{shotton20137scenes}.}
We report the average of the per-scene median translation and rotation errors.}
\centering
\small
\setlength{\tabcolsep}{6pt}
\begin{tabular}{lccccc}
    \toprule
    & & \multicolumn{2}{c}{Cambridge Landmarks}
      & \multicolumn{2}{c}{7Scenes} \\
    \cmidrule(lr){3-4} \cmidrule(lr){5-6}
    Method
    & Pose Refinement
    & Size (MB)
    & Avg. (cm/$^\circ$)
    & Size (MB)
    & Avg. (cm/$^\circ$) \\
    \midrule
    \textbf{SplitGS-Loc~(Ours)}
    & & 97 & 8.7/0.15 & 88 & 1.0/0.26 \\
    \quad + Feature Rendering~\cite{huang2025stdloc}
    & \checkmark & 415 & \textbf{8.4/0.13} & 888 & 0.9/0.23 \\
    \quad + GS-CPR~\cite{liu2025gscpr}
    & \checkmark & 177 & 14.9/0.25 & 263 & \textbf{0.6/0.19} \\
    \bottomrule
\end{tabular}
\label{tab:suppl_refinement}
\end{table}
\section{Robustness under Appearance Changes}
\label{sec:ecl}

We further evaluate SplitGS-Loc under appearance changes using Extended Cambridge Landmarks (ECL)~\cite{ferens2025hyperpose}.
ECL provides summer, winter, and evening variants of the Cambridge Landmarks test images while preserving their original camera poses.
For both STDLoc~\cite{huang2025stdloc} and SplitGS-Loc, we use the same maps as in the Cambridge Landmarks evaluation and directly evaluate them on ECL without adaptation.
Tab.~\ref{tab:ecl} shows that SplitGS-Loc consistently outperforms STDLoc across different appearance conditions.
Since both methods use the same image feature extractor and original training views, the results highlight the robustness of our localization-specialized GSFFs construction to unseen appearance changes.
Notably, SplitGS-Loc achieves this without adaptation to ECL, while retaining its efficient mapping process.

\begin{table}[t]
\caption{\textbf{Localization results on Extended Cambridge Landmarks~\cite{ferens2025hyperpose}.}
We report the average of the per-scene median translation and rotation errors.}
\centering
\small
\setlength{\tabcolsep}{8pt}
\begin{tabular}{lcccc}
    \toprule
    Method
    & \makecell{Mapping\\Time}
    & \makecell{Summer\\(cm/$^\circ$)}
    & \makecell{Winter\\(cm/$^\circ$)}
    & \makecell{Evening\\(cm/$^\circ$)} \\
    \midrule
    STDLoc~\cite{huang2025stdloc}
    & 2h & \textbf{19.5}/0.34 & 20.1/0.36 & 21.1/0.37 \\
    \textbf{SplitGS-Loc~(Ours)}
    & 7min & \textbf{19.5/0.27} & \textbf{19.5/0.28} & \textbf{20.2/0.31} \\
    \bottomrule
\end{tabular}
\label{tab:ecl}
\end{table}

\newpage

\end{document}